\documentclass[sigconf,nonacm]{acmart}

\AtBeginDocument{%
  }

\usepackage{pifont}
\newcommand{\cmark}{\ding{51}}
\newcommand{\xmark}{\ding{55}}

\usepackage{multirow}
\usepackage{amsthm}
\usepackage{enumitem}

\theoremstyle{remark}
\newtheorem{remark}{Remark}
\newtheorem{assumption}{Assumption}

\begin{document}

\title{LAVA: Layered Audio-Visual Anti-tampering Watermarking for Robust Deepfake Detection and Localization}

\author{Bokang Zeng}
\email{bokang.zeng@student.unsw.edu.au}
\affiliation{
  \department{School of Computer Science and Engineering}
  \institution{UNSW Sydney}
  \city{Sydney}
  \country{Australia}
}

\author{Zheng Gao}
\email{zheng.gao1@unsw.edu.au}
\affiliation{
  \department{School of Computer Science and Engineering}
  \institution{UNSW Sydney}
  \city{Sydney}
  \country{Australia}
}

\author{Xiaoyu Li}
\email{xiaoyu.li2@student.unsw.edu.au}
\affiliation{
  \department{School of Computer Science and Engineering}
  \institution{UNSW Sydney}
  \city{Sydney}
  \country{Australia}
}

\author{Xiaoyan Feng}
\email{xiaoyan.feng@griffithuni.edu.au}
\affiliation{
  \department{School of Information and Communication Technology} 
  \institution{Griffith University}
  \city{Brisbane}
  \country{Australia}
}

\author{Jiaojiao Jiang}
\email{jiaojiao.jiang@unsw.edu.au}
\affiliation{
  \department{School of Computer Science and Engineering}
  \institution{UNSW Sydney}
  \city{Sydney}
  \country{Australia}
}

\begin{abstract}
Proactive watermarking offers a promising approach for deepfake tamper detection and localization in short-form videos. However, existing methods often decouple audio and visual evidence and assume that watermark signals remain reliable under real-world degradations, which makes tamper localization vulnerable to multimodal misalignment and compression distortions. Moreover, existing semi-fragile visual watermarking methods often degrade significantly under codec compression because their embedding bands overlap with compression-sensitive frequency regions. To address these limitations, we propose \underline{\textbf{L}}ayered \underline{\textbf{A}}udio-\underline{\textbf{V}}isual \underline{\textbf{A}}nti-tampering Watermarking (\textbf{LAVA}), a calibration-aware audio--visual watermark fusion framework for deepfake tamper detection and localization. LAVA leverages cross-modal watermark fusion and calibration-aware alignment to preserve consistent and reliable tamper evidence under compression and audio-visual asynchrony, thereby enabling robust tamper localization. Extensive experiments demonstrate that LAVA achieves near-perfect detection performance (AP = 0.999), remains robust to compression and multimodal misalignment, and significantly improves tamper localization reliability over existing audio-visual fusion baselines.


\end{abstract}



\maketitle

\section{Introduction}

The rapid growth of short-form video platforms has dramatically lowered the barrier to creating and disseminating audio-visual deepfakes. Modern face-swapping, lip-sync, and voice-cloning systems can now generate highly realistic forged videos within seconds, and such content often remains convincing even after platform-side compression, transcoding, and re-packaging \cite{10.1145/3394171.3413532,10.1145/3664647.3680795,CAI2023103818,10.1145/3801962}. As a result, the central challenge is no longer merely to determine whether a video has been manipulated, but also to localize \emph{when} and \emph{where} the manipulation occurs. This capability is critical for platform moderation, media provenance, and forensic verification.

Existing deepfake detection approaches can be broadly divided into \emph{passive} forensics and \emph{proactive} authentication. Passive methods detect manipulations by mining statistical artifacts left by the generation process, such as facial boundary inconsistencies, frequency anomalies, temporal incoherence, or cross-modal mismatch \cite{Rossler_2019_ICCV++,Li_2020_CVPR,10.1007/978-3-030-58610-2_6,Zheng_2021_ICCV,Wang_2023_CVPR,Yan_2023_ICCV,10.1145/3394171.3413700,Zhou_2021_ICCV}. Although these methods have achieved strong performance on standard benchmarks, they fundamentally rely on artifacts that become weaker as generative models improve, and are often fragile under post-processing such as compression and transcoding \cite{Haliassos_2022_CVPR,10.1145/3801962}. In contrast, proactive watermarking embeds a verifiable signal before distribution and checks its integrity at retrieval time, offering a more controllable path toward deepfake detection and localization \cite{lin1999review,10.1145/3640466}.

However, existing watermark-based authentication schemes remain insufficient for realistic short-video deployment. On the one hand, robust watermarks are designed to survive aggressive compression and common editing operations, making them suitable for copyright verification, but they are often not sufficiently sensitive to localized tampering. On the other hand, fragile or semi-fragile watermarks can react to content modification, yet may collapse under benign platform distortions \cite{Zhu_2018_ECCV,Tancik_2020_CVPR,10.1145/3474085.3475324,10.1145/3640466}. For deepfake detection, these two requirements must be satisfied simultaneously: the integrity signal should remain verifiable under benign compression, while still being sensitive to localized face manipulation, voice cloning, or joint audio-visual forgery. In other words, the goal is not merely to preserve a watermark, but to preserve \emph{interpretable}, \emph{localizable}, and \emph{calibrated} tamper evidence under realistic distribution pipelines.

A more fundamental limitation is that existing proactive authentication methods are largely \emph{unimodal}. Visual methods focus on recovering or localizing image/video watermarks, such as WAM, EditGuard, OmniGuard, and VideoSeal \cite{sander2025watermark,Zhang_2024_CVPR,Zhang_2025_CVPR,fernandez2024videosealopenefficient}, while audio methods target waveform-level watermark detection, such as AudioSeal and WavMark \cite{10.5555/3692070.3693829,chen2024wavmarkwatermarkingaudiogeneration}. Although these approaches are effective within their respective modalities, they typically model audio and visual evidence independently, or combine them with simple score averaging, implicitly assuming that both modalities remain synchronized, stable, and equally reliable at inference time. This assumption is often violated in real short-video platforms.

In practice, the audio and visual streams of an uploaded video undergo different forms of degradation. The video track may be re-encoded with lossy codecs such as H.264, while the audio track may be transcoded into AAC or MP3; container remuxing can further introduce sub-second audio-visual offsets, and some platforms may even apply slight tempo changes. Under such distortions, audio and visual watermarks exhibit markedly different failure modes: the visual watermark may collapse globally under strong compression while the audio watermark remains usable; conversely, the audio watermark may degrade under temporal stretching or asynchrony while the visual channel still carries valid evidence. Consequently, the real challenge is not simply how to combine two scores, but how to make reliable tamper decisions when modalities are misaligned, channel reliability is asymmetric, and failure patterns differ across modalities.

In this paper, we argue that the root cause of this failure is an overlooked assumption implicitly made by existing audio-visual watermarking pipelines: a \emph{synchronous reliability assumption}, namely, that all modalities provide stable and directly comparable integrity signals at the same time. Our key observation is that audio and visual watermarks exhibit a complementary form of \emph{robustness asymmetry} under deployment distortions: when one modality fails, the other often remains informative. Therefore, multimodal watermark-based deepfake detection should not be formulated as a simple average over two always-reliable channels, but as a layered inference problem over heterogeneous integrity signals whose reliability varies with time and distortion type.

Motivated by this observation, we propose \textbf{LAVA} (\textbf{L}ayered \textbf{A}udio \\ \textbf{V}isual \textbf{A}nti-tampering Watermarking), a layered audio-visual watermark fusion framework for deepfake detection and spatio-tem\-po\-ral tamper localization. LAVA embeds two independent semi-fragile watermarks into the visual and audio modalities, and organizes inference around a simple principle: restore alignment first, identify channel failure second, and perform local fusion and probability calibration last. Concretely, LAVA first corrects temporal stretching to recover audio-visual alignment; it then applies a reliability gate to determine whether the visual channel has globally failed under compression; when both channels are usable, it performs frame-wise confidence-weighted fusion; finally, it calibrates the fused scores into interpretable tamper probabilities using temperature scaling \cite{8269806,JAIN20052270,pmlr-v70-guo17a}. This design enables robust detection under compression and asynchrony, while producing spatial tamper maps only when the visual channel is sufficiently reliable.

Extensive experiments on LAV-DF and multiple deployment distortion settings demonstrate that LAVA achieves consistently strong performance across clean, compressed, misaligned, and time-stretched conditions \cite{CAI2023103818}. Beyond improving overall detection accuracy, LAVA substantially improves temporal localization stability and calibration quality over unimodal baselines and naive fusion. These results show that the advantage of LAVA does not come merely from using an additional modality, but from explicitly modeling how multimodal watermark evidence fails in realistic deployment environments.

In summary, our main contributions are as follows:
\begin{itemize}
    \item We identify a key limitation of existing audio-visual water\-mark-based detection pipelines: they implicitly rely on a synchronous reliability assumption that often breaks under compression, transcoding, temporal offset, and time stretching.

    \item We propose LAVA, a layered audio-visual watermark fusion framework that systematically addresses three core challenges in realistic deployment: modality misalignment, channel failure, and score miscalibration.

    \item We reveal and exploit the robustness asymmetry between audio and visual watermarks under deployment distortions, turning it into a reliable cross-modal integrity inference mechanism that maintains detection capability when at least one modality remains informative.

    \item Experiments on LAV-DF under diverse compression, offset, stretching, and platform-simulation settings show that LAVA consistently outperforms unimodal baselines and naive fusion in detection accuracy, temporal localization stability, and calibration quality.
\end{itemize}
\section{Related Work}
\label{sec:related}

\subsection{Passive Deepfake Detection and Localisation}
\label{sec:rw:passive}

Passive deepfake detection learns to identify synthesis artefacts from data, typically without modifying the original content.
Early visual detectors focus on spatial cues: the FaceForensics++ benchmark~\cite{Rossler_2019_ICCV++}
 standardised evaluation across four manipulation types; Face X-Ray~\cite{Li_2020_CVPR} exploits blending boundaries shared by most face-swap pipelines; F3-Net~\cite{10.1007/978-3-030-58610-2_6} shifts detection to the frequency domain via DCT decomposition, improving robustness under compression.
Generalisation remains the central challenge.
Forgery-agnostic augmentation~\cite{Shiohara_2022_CVPR} and common-feature disentanglement~\cite{Yan_2023_ICCV} reduce overfitting to method-specific patterns, while temporal coherence modeling~\cite{Zheng_2021_ICCV} and alternating spatial--temporal freezing~\cite{Wang_2023_CVPR} extend detection to the video setting.
In the audio domain, ASVspoof~\cite{WANG2020101114} established standard benchmarks; countermeasures have since progressed from end-to-end waveform models (RawNet2~\cite{9414234}) through spectro-temporal graph attention (AASIST~\cite{9747766}) to self-supervised wav2vec fine-tuning~\cite{tak2022automaticspeakerverificationspoofing}.
Audio-visual methods exploit cross-mod\-al inconsistency---modality dissonance scoring~\cite{10.1145/3394171.3413700}, joint stream classification~\cite{Zhou_2021_ICCV}, lip-motion self-supervision~\cite{Haliassos_2022_CVPR}---and have recently scaled to temporal localisation on the LAV-DF benchmark~\cite{CAI2023103818,10.1145/3664647.3680795}.
Across all three modality settings, these detectors remain fundamentally limited by their dependence on artefact patterns that improve with generator quality and degrade under platform recompression.
LAVA circumvents this dependency by detecting the absence of a proactively embedded watermark (\S\ref{sec:exp:attr}).

\begin{figure*}[!t]
\centering
\includegraphics[width=\textwidth]{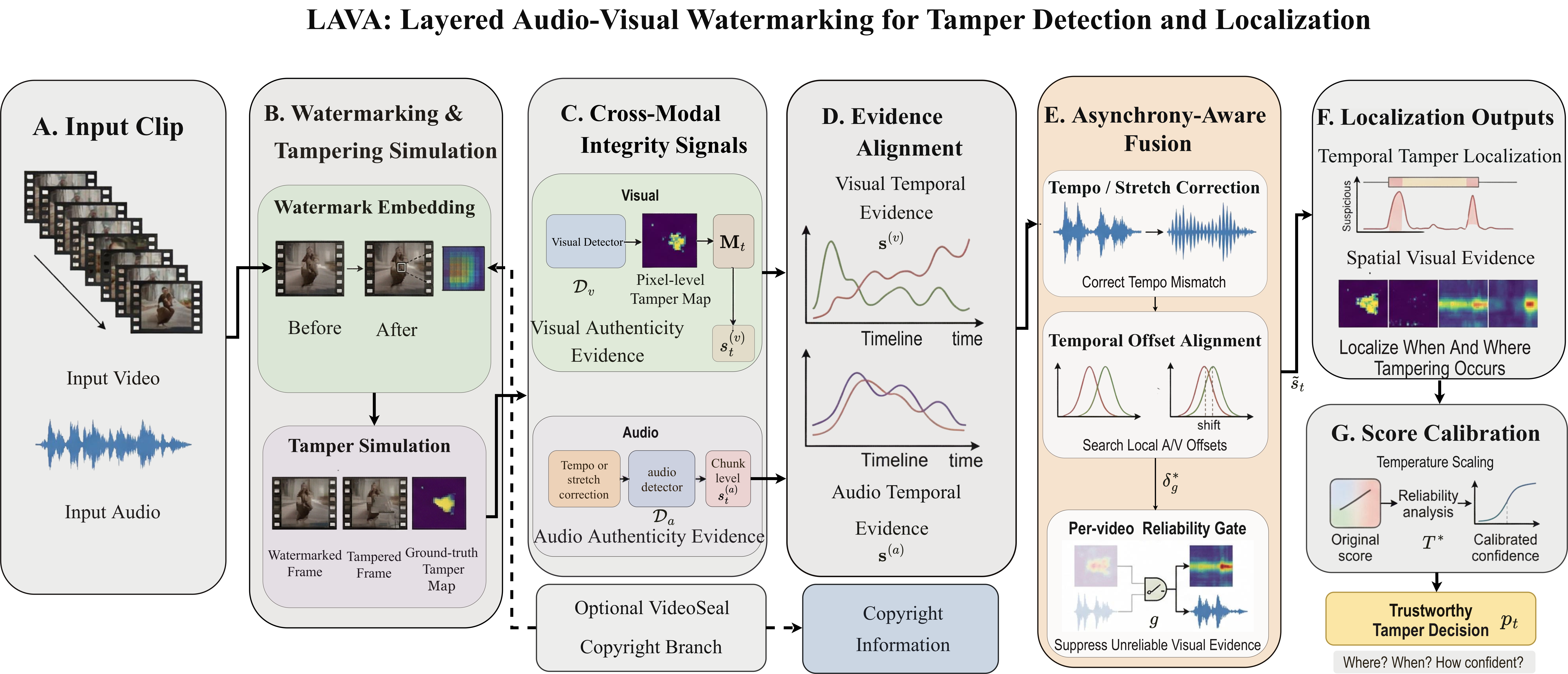}
\Description{LAVA pipeline diagram showing input video split into visual and audio tracks, watermark embedding, channel distortions, watermark detection producing per-frame scores, and the four-layer fusion pipeline outputting calibrated tamper probabilities and spatial maps.}
\caption{Overview of the LAVA pipeline. Independent visual and audio
integrity watermarks are embedded before distribution. At inference,
per-frame detector scores pass through a four-layer hierarchical
process---temporal restoration, reliability gate, confidence-weighted
fusion, and calibration---yielding a calibrated per-frame tamper
probability~$p_t$ and, when the visual channel is reliable, a
pixel-level tamper map~$\mathbf{M}_t$.}
\label{fig:pipeline}
\end{figure*}

\subsection{Proactive Watermark-Based Media Authentication}
\label{sec:rw:watermark}

Proactive integrity verification embeds a signal into authentic content before distribution and checks its survival at inspection time.
Classical fragile watermarks~\cite{lin1999review} enable bit-exact authentication but cannot tolerate any benign processing; semi-fragile designs relax this constraint, as recently demonstrated by FaceSigns~\cite{10.1145/3640466} with end-to-end neural watermarking for deepfake countering.

Learned watermarking has substantially advanced the robust\-ness--imperceptibility trade-off since the introduction of HiDD\-eN~\cite{Zhu_2018_ECCV}, which pioneered encoder--decoder training with differentiable noise layers.
Physical-world augmentation (StegaStamp~\cite{Tancik_2020_CVPR}), mini-batch compression simulation (MBRS~\cite{10.1145/3474085.3475324}), and diffusion-latent embedding (Stable Signature~\cite{Fernandez_2023_ICCV}, Tree-Ring~\cite{NEURIPS2023_b54d1757}) have progressively widened the robustness envelope.
For tamper localisation specifically, WAM~\cite{sander2025watermark} provides pixel-level maps via JND masking, while EditGuard~\cite{Zhang_2024_CVPR} and OmniGuard~\cite{Zhang_2025_CVPR} combine localisation with copyright protection through dual watermark layers.

These image-level methods have been extended independently to other modalities: VideoSeal~\cite{fernandez2024videosealopenefficient} adds temporal propagation for video; AudioSeal~\cite{10.5555/3692070.3693829} and WavMark~\cite{chen2024wavmarkwatermarkingaudiogeneration} embed localised watermarks in the audio waveform.
Yet each line of work remains strictly unimodal---none addresses how to jointly interpret audio and visual watermark evidence when deployment distortions selectively destroy one channel.

\subsection{Cross-Modal Fusion and Confidence Calibration}
\label{sec:rw:fusion}

Score-level fusion of heterogeneous detectors has been extensively studied in biometric verification, where normalisation and weighted combination of match scores from face, fingerprint, and iris modalities yield consistent accuracy gains~\cite{JAIN20052270}.
The broader multimodal learning literature~\cite{8269806} categorises fusion strategies into early, late, and hybrid approaches, with late (score-level) fusion being the most modular and widely adopted when subsystems are independently trained.
Post-hoc confidence calibration further improves the reliability of fused outputs, with approaches ranging from Bayesian binning~\cite{naeini2015bbq} to temperature scaling~\cite{pmlr-v70-guo17a}, the latter being the simplest and most effective single-parameter method.

A common assumption underlying these approaches is that all input channels carry discriminative information simultaneously.
This assumption fails in the watermark fusion setting: aggressive H.264 compression can collapse the visual watermark globally while leaving the audio watermark intact, and conversely, tempo adjustment can destroy the audio watermark while the visual channel survives.
LAVA addresses this \emph{selective channel failure} with hierarchical reliability-aware inference---per-video gating, per-frame confidence weighting, and ECE-minimising calibration---and is, to our knowledge, the first framework to fuse audio and visual watermark evidence for joint tamper detection and localisation.

\section{Method}\label{sec:method}

We present \textbf{LAVA} (\textbf{L}ayered \textbf{A}udio-\textbf{V}isual
\textbf{A}nti-tampering), a framework for video deepfake detection and
temporal--spatial tamper localisation based on independent visual and audio
semi-fragile watermarks.
Figure~\ref{fig:pipeline} provides an overview (the optional VideoSeal copyright branch shown therein is an auxiliary provenance channel not used in the experiments).
The key insight is that any localised tampering, including face swap, voice
cloning, and joint manipulation, necessarily destroys the watermark in the
affected modality and time interval.
LAVA exploits this through \textbf{hierarchical reliability-aware inference}:
temporal alignment is restored before cross-modal fusion, global channel
failures are resolved before local ones, and raw scores are calibrated into
tamper probabilities.
We formalise the setup in~\S\ref{sec:formulation}.

\subsection{Problem Formulation}\label{sec:formulation}

\begin{definition}[Watermarked Video and Integrity Scores]
\label{def:video}
Let $\mathbf{x}^w = (\mathbf{V}, \mathbf{A})$ denote a watermarked video
with visual track $\mathbf{V} = \{\mathbf{I}_t\}_{t=1}^{T}$ at frame
rate~$f$ and audio track~$\mathbf{A} \in \mathbb{R}^{N}$.
Prior to distribution, independent semi-fragile watermarks are embedded via
a per-frame visual encoder~$\mathcal{E}_v$ and a waveform-level audio
encoder~$\mathcal{E}_a$.
At inference, the associated detectors $\mathcal{D}_v$, $\mathcal{D}_a$
extract per-frame \emph{integrity scores} under tamper polarity
(high\,$=$\,watermark absent\,$=$\,likely tampered):
\begin{align}
  \mathcal{D}_v(\mathbf{I}_t) &= \mathbf{M}_t \in [0,1]^{H \times W},
  &s_t^{(v)} &= \tfrac{1}{HW}\sum\nolimits_{h,w}
    \mathbf{M}_t[h,w], \label{eq:vis} \\[2pt]
  \mathcal{D}_a(\mathbf{A}) &= \mathbf{d}^{(a)} \in [0,1]^{N},
  &s_t^{(a)} &= 1 - \tfrac{1}{|\mathcal{C}_t|}\sum\nolimits_{n \in \mathcal{C}_t}
    d_n^{(a)}, \label{eq:aud}
\end{align}
where $\mathbf{M}_t$ is the pixel-level tamper map, $\mathbf{d}^{(a)}$ is
the per-sample watermark-presence vector (high\,$=$\,watermark detected),
and $\mathcal{C}_t = \{n : (t{-}1)/f \le n/f_s < t/f\}$ is the
audio-sample index set aligned to frame~$t$ at sampling rate~$f_s$.
The complement in Eq.~\eqref{eq:aud} converts watermark presence into
tamper polarity.
We write $\mathbf{s}^{(v)}, \mathbf{s}^{(a)} \in [0,1]^T$ for the two
score sequences.
\end{definition}

\begin{assumption}[Deployment Distortion Model]
\label{asm:distortion}
During distribution, the watermarked video $\mathbf{x}^w$ may undergo
platform-induced distortions.
We model their effect at the score-sequence level as follows (this is a
first-order approximation; the actual impact on scores depends on the
detector behaviour):
\begin{enumerate}[label=(\roman*),nosep,leftmargin=*]
  \item \textbf{Temporal stretch.}
    The audio track is resampled by factor $\alpha \neq 1$ while the video
    frame rate is preserved:
    $s_t^{(a)} \mapsto s_{\lfloor \alpha t \rfloor}^{(a)}$,\;
    $s_t^{(v)} \mapsto s_t^{(v)}$;
  \item \textbf{Lossy compression.}
    Aggressive video codecs destroy the visual watermark globally while
    leaving the audio channel intact:
    $s_t^{(v)} \to 1\;\;\forall\, t$,\;
    $s_t^{(a)} \mapsto s_t^{(a)}$;
  \item \textbf{A/V offset.}
    Container remuxing introduces a temporal shift $\delta$\,s between the
    two tracks:
    $s_t^{(a)} \mapsto s_{t - \lfloor \delta f \rceil}^{(a)}$,\;
    $s_t^{(v)} \mapsto s_t^{(v)}$.
\end{enumerate}
Each individual distortion typically affects one modality more severely than the other; in practice, multiple distortions may co-occur (\S\ref{sec:exp:robustness}).
\end{assumption}

\begin{definition}[Detection Objective]
\label{def:objective}
Given a query video $\mathbf{x}^q$ that has potentially undergone both
tampering and the distortions in Assumption~\ref{asm:distortion}, the
detector $\mathcal{F}$ produces per-frame calibrated tamper
probabilities~$p_t$ and, when available, pixel-level tamper maps:
\begin{align}
  \mathcal{F}(\mathbf{x}^q)
  &= \bigl\{\,(p_t,\; (1-g) \cdot \mathbf{M}_t)\,\bigr\}_{t=1}^{T},
  \qquad
  p_t \in [0,1],\;\;
  g \in \{0,1\}, \label{eq:objective}
\end{align}
where $p_t$ satisfies
$\mathbb{P}(y_t{=}1 \mid p_t{=}p) \approx p$
with $y_t \in \{0,1\}$ the ground-truth tamper label, and
$g$ is a video-level visual-channel failure indicator:
$g{=}1$ signals that the visual watermark is globally compromised,
suppressing the spatial map~$\mathbf{M}_t$;
$g{=}0$ indicates the visual channel is reliable.
\end{definition}

The score sequences $\mathbf{s}^{(v)},\,\mathbf{s}^{(a)} \in [0,1]^T$
cannot be naively combined: under Assumption~\ref{asm:distortion}, one or
both channels may produce uniformly elevated scores on authentic frames,
and the failure patterns are modality-dependent and structurally distinct
from genuine tampering.
LAVA addresses this through the layered pipeline described in
\S\ref{sec:signals}--\ref{sec:calibration}.

\subsection{Cross-Modal Integrity Signals}\label{sec:signals}

The two detectors $\mathcal{D}_v$, $\mathcal{D}_a$ defined in
Definition~\ref{def:video} operate in complementary domains:
$\mathcal{D}_v$ in compression-sensitive frequency bands,
$\mathcal{D}_a$ in the waveform time domain.
LAVA makes no assumption on their internal architecture and requires
only the interface specified by Eqs.~\eqref{eq:vis}--\eqref{eq:aud}.

\begin{remark}[Cross-Modal Robustness Asymmetry]
\label{rem:asymmetry}
Let $\mathcal{D}_{\mathrm{fail}}^{(v)}$ and
$\mathcal{D}_{\mathrm{fail}}^{(a)}$ denote the sets of distortion types
that cause catastrophic failure of each channel.
Under the distortions considered in Assumption~\ref{asm:distortion},
$\mathcal{D}_{\mathrm{fail}}^{(v)} \cap
 \mathcal{D}_{\mathrm{fail}}^{(a)} = \varnothing$:
lossy video compression collapses $\mathbf{s}^{(v)}$ while preserving
$\mathbf{s}^{(a)}$, and temporal stretch degrades $\mathbf{s}^{(a)}$
while preserving $\mathbf{s}^{(v)}$.
This asymmetry ensures that at least one channel remains informative for
each deployment distortion, motivating cross-modal fusion.
A direct corollary is that the survival pattern reveals the attack type:
face swap ($s_t^{(v)}\!\uparrow,\; s_t^{(a)}\!\downarrow$),
voice cloning ($s_t^{(v)}\!\downarrow,\; s_t^{(a)}\!\uparrow$),
or joint deepfake ($s_t^{(v)}\!\uparrow,\; s_t^{(a)}\!\uparrow$),
without any additional classifier.
\end{remark}

\paragraph{Temporal stretch correction.}
Before fusion, LAVA corrects the misalignment from
Assumption~\ref{asm:distortion}\,(i).
The stretch factor $\hat{\alpha} = L_{\mathrm{audio}} / (T/f)$ is
directly observable; when $|\hat{\alpha} - 1| > 0.01$, the audio is
restored via
$\mathbf{A}_{\mathrm{corr}} = \mathcal{R}_{1/\hat{\alpha}}(\mathbf{A})$,
where $\mathcal{R}_{\beta}$ is the linear-interpolation resampling operator
at rate ratio~$\beta$, chosen for its approximate self-inverse property
$\mathcal{R}_{1/\alpha} \circ \mathcal{R}_{\alpha} \approx \mathrm{Id}$.

\subsection{Asynchrony-Aware Fusion}\label{sec:fusion}

After temporal correction, the score sequences
$\mathbf{s}^{(v)},\,\mathbf{s}^{(a)} \in [0,1]^T$ must be combined into a
single fused signal.
Two structurally distinct reliability failures arise in practice:
catastrophic collapse (aggressive compression destroys the visual watermark
globally) and local degradation (one modality becomes unstable at specific
frames).
LAVA handles them at two granularities: a per-video hard gate resolves
global failure, and a per-frame soft weighting addresses local instability,
ensuring that global failure modes are resolved before local fusion is
attempted.

\paragraph{Per-video reliability gate.}
Under strong compression, $s_t^{(v)}$ collapses to uniformly high values
across all frames, eliminating discriminative power.
The key observation is that this failure mode is \emph{temporally uniform},
whereas genuine tampering is \emph{temporally sparse} (only the manipulated
interval shows elevated scores).
The global visual mean therefore serves as a reliable proxy for channel
failure:
\begin{equation}\label{eq:gate}
  g = \mathbf{1}\!\Bigl[\,\bar{s}^{(v)} > \tau\,\Bigr],
  \qquad
  \bar{s}^{(v)} = \frac{1}{T}\sum_{t=1}^{T} s_t^{(v)},
\end{equation}
where $\tau = 0.1$ is selected via grid search.
When $g = 1$, the visual channel is suppressed for the entire video and
$\mathbf{M}_t$ is nulled.
This gate targets the localised-tampering regime; if the majority of frames
are tampered, the mean-based test may fail to distinguish tampering from
compression (\S\ref{sec:exp:ablation}).

\paragraph{Temporal offset alignment.}
Let $s_{\delta,t}^{(a)} := s_{t-\lfloor\delta f\rceil}^{(a)}$ denote the
audio score shifted by $\delta$\,s, and let
$\mathcal{O} = \{-1.0,\, -0.75,\, \ldots,\, +1.0\}$\,s be the candidate
offset set.
Define the fused sequence under offset~$\delta$ as
\begin{equation}\label{eq:fused-delta}
  \hat{\mathbf{s}}_g(\delta) =
  \begin{cases}
    \{s_{\delta,t}^{(a)}\}_{t=1}^T, & g=1,\\[3pt]
    \{\hat{s}_t(\delta)\}_{t=1}^T,   & g=0,
  \end{cases}
\end{equation}
where $\hat{s}_t(\delta)$ is the confidence-weighted combination in
Eq.~\eqref{eq:fused} evaluated at offset~$\delta$.
The optimal offset is
\begin{equation}\label{eq:offset-sel}
  \delta^{*}_g \in \bigl\{\delta \in \mathcal{O} :
    \mathcal{J}\bigl(\mathbf{y},\,\hat{\mathbf{s}}_g(\delta)\bigr)
    = \max_{\delta' \in \mathcal{O}}\,
    \mathcal{J}\bigl(\mathbf{y},\,\hat{\mathbf{s}}_g(\delta')\bigr)
  \bigr\},
\end{equation}
where $\mathcal{J}$ is a frame-level detection performance measure
(instantiated as average precision in experiments), with ties broken by
smallest $|\delta|$.
In the current evaluation this selection is oracle (i.e.\ uses test-set
labels), establishing the performance ceiling of offset alignment; at
deployment it can be replaced by unsupervised cross-correlation
(\S\ref{sec:exp:robustness}).

\begin{table*}[!htbp]
\centering
\caption{Detection performance across three datasets under clean and JPEG compression. LAV-DF and VoxCeleb2 report frame-level AP; FakeAVCeleb reports video-level AUC. VoxCeleb2 averaged over five attack types with async 0.65\,s A/V tamper. Best in \textbf{bold}, second-best \underline{underlined}.}
\label{tab:main}
\small
\begin{tabular}{l cc cc cc cc}
\toprule
 & \multicolumn{2}{c}{LAV-DF (AP)} & \multicolumn{2}{c}{LAV-DF (IoU)} & \multicolumn{2}{c}{FakeAVCeleb (AUC)} & \multicolumn{2}{c}{VoxCeleb2 (AP)} \\
\cmidrule(lr){2-3} \cmidrule(lr){4-5} \cmidrule(lr){6-7} \cmidrule(lr){8-9}
Method & Clean & JPEG & Clean & JPEG & Clean & JPEG & Clean & JPEG \\
\midrule
Visual-only       & \textbf{1.000} & 0.453  & ---   & ---   & 0.833  & 0.565  & 0.690  & 0.214  \\
Audio-only        & 0.999           & \textbf{0.999}  & ---   & ---   & 0.814  & \underline{0.809}  & 0.781  & \textbf{0.780}  \\
Na\"{\i}ve fusion & \textbf{1.000} & \underline{0.999}  & \underline{0.950} & 0.750 & \underline{0.958}  & 0.792  & \underline{0.953}  & 0.724  \\
\midrule
LAVA (ours)       & \textbf{1.000} & \textbf{0.999} & \textbf{0.952} & \textbf{0.955} & \textbf{1.000} & \textbf{0.810} & \textbf{0.991} & \underline{0.773} \\
\bottomrule
\end{tabular}
\end{table*}

\paragraph{Per-frame confidence-weighted fusion ($g{=}0$).}
Watermark evidence is piecewise smooth within authentic or tampered
segments; abrupt local fluctuations indicate detector instability rather
than semantic evidence.
Since $s_t^{(m)} \in [0,1]$, the sample variance satisfies
$\sigma^2_{W}(s^{(m)},t) \le 1/4$.
We therefore define per-frame confidence by normalising the local variance
to $[0,1]$:
\begin{equation}\label{eq:confidence}
  c_t^{(m)}
  = \bigl(1 - 4\,\sigma^2_{W}(s^{(m)},\, t)\bigr)_{\!+}\,,
  \quad m \in \{a, v\},
\end{equation}
where $\sigma^2_{W}(s^{(m)},\, t)$ is the sample variance of
$\{s_k^{(m)}\}_{k=t-W}^{t+W}$ with half-window $W{=}3$.
When $\sigma^2_W = 1/4$ (maximal instability), $c_t^{(m)} = 0$;
when $\sigma^2_W = 0$ (perfect stability), $c_t^{(m)} = 1$.
The fused score is
\begin{equation}\label{eq:fused}
  \hat{s}_t =
  \begin{cases}
    \displaystyle
    \frac{c_t^{(a)}\, s_{\delta^*_0,\,t}^{(a)}
         \;+\; c_t^{(v)}\, s_t^{(v)}}
         {c_t^{(a)} + c_t^{(v)}},
    & c_t^{(a)} + c_t^{(v)} > 0, \\[10pt]
    \tfrac{1}{2}\bigl(s_{\delta^*_0,\,t}^{(a)} + s_t^{(v)}\bigr),
    & c_t^{(a)} + c_t^{(v)} = 0.
  \end{cases}
\end{equation}
The gate-unified output combines both branches:
\begin{equation}\label{eq:gate-override}
  \tilde{s}_t =
  \begin{cases}
    s_{\delta^*_1,\,t}^{(a)}, & g=1,\\[3pt]
    \hat{s}_t,                 & g=0.
  \end{cases}
\end{equation}

\subsection{Score Calibration}\label{sec:calibration}

The fused scores $\tilde{s}_t$ are not inherently calibrated: a score of
$0.7$ does not correspond to a $70\%$ tampering probability, because the
upstream watermark detectors were not trained with a calibration objective.
To satisfy the calibration requirement
$\mathbb{P}(y_t{=}1 \mid p_t{=}p) \approx p$ in
Definition~\ref{def:objective}, we apply temperature scaling~\cite{pmlr-v70-guo17a}: a single
scalar $T > 0$ rescales the logits to minimise the expected calibration
error (ECE):
\begin{equation}\label{eq:tempscale}
  T^* = \operatorname*{arg\,min}_{T > 0}\;
    \mathrm{ECE}\bigl(\mathbf{y},\;\sigma(\boldsymbol{\ell}/T)\bigr),
\end{equation}
where $\ell_t = \mathrm{logit}(\tilde{s}_t)$, $\sigma$ is the sigmoid
function, and ECE uses $B{=}10$ equal-width bins.
In practice, $T^*$ is found by grid search over $300$ log-spaced candidates
in $[10^{-2},\, 10]$.
The calibrated tamper probability is
\begin{equation}\label{eq:calibrated}
  p_t = \sigma\bigl(\ell_t \,/\, T^*\bigr).
\end{equation}
We minimise ECE rather than NLL because ECE directly measures the gap
between predicted confidence and observed accuracy.
Temperature scaling preserves score ranking (and hence AP) while
substantially reducing calibration error (\S\ref{sec:exp:ablation}).

\paragraph{Inference summary.}
The four stages run sequentially: temporal stretch correction,
reliability gate (Eq.~\ref{eq:gate}),
offset alignment and confidence-weighted fusion
(Eqs.~\ref{eq:offset-sel}--\ref{eq:gate-override}),
and temperature calibration
(Eqs.~\ref{eq:tempscale}--\ref{eq:calibrated}).
The output is a per-frame calibrated tamper probability~$p_t$ and, when
$g{=}0$, a pixel-level map~$\mathbf{M}_t$; when $g{=}1$, LAVA abstains
from spatial localisation.
All post-extraction stages are signal-processing operations with no
learnable parameters beyond~$T^*$, adding ${<}\,55$\,ms combined overhead.

\section{Experiments}
\label{sec:exp}

We evaluate LAVA from three aspects: \emph{(i)} whether it improves detection, localisation, and calibration over unimodal baselines and simple fusion; \emph{(ii)} which components of the layered design are responsible for these gains; and \emph{(iii)} how it performs under deployment distortions, including compression, temporal offset, tempo perturbation, and joint cross-modal degradation.

\subsection{Experimental Setup}
\label{sec:exp:setup}

\textbf{Datasets.}
We evaluate LAVA on three complementary benchmarks.
\textbf{LAV-DF}~\cite{CAI2023103818} contains 500 groups with frame-level temporal annotations covering face swap, voice conversion, and joint audio-visual manipulation.
\textbf{FakeAVCeleb}~\cite{khalid2022fakeavcelebnovelaudiovideomultimodal} contains 2{,}000 videos from four manipulation categories with video-level labels.
\textbf{VoxCeleb2}~\cite{Chung_2018} is used to construct a controlled benchmark of 200 real videos with synthetic 0.65\,s asynchronous audio-visual tampering and five visual attack types, including blur, black-box occlusion, mosaic, noise injection, and white-out.
Unless otherwise specified, LAV-DF is used as the primary benchmark for temporal detection, localisation, and calibration, FakeAVCeleb for video-level generalisation, and VoxCeleb2 for controlled analysis of asynchronous tampering and watermark quality.

\textbf{Distortion settings.}
All three datasets are evaluated under \textbf{Clean} (no re-encoding) and \textbf{JPEG} $q{=}23$ + MP3 128\,k.
For LAV-DF, we further consider \textbf{H.264 CRF\,23} and \textbf{H.264 CRF\,28} + MP3 128\,k to simulate progressively stronger platform-style compression.

\textbf{Compared methods.}
We compare LAVA with three internal baselines: \textbf{Visual-only}, which uses only the visual watermark branch; \textbf{Audio-only}, which uses only the audio watermark branch; and \textbf{Na\"{\i}ve fusion}, defined as $\hat{s}_t = 0.5\,s_t^{(v)} + 0.5\,s_t^{(a)}$ without temporal alignment, reliability gating, or calibration.
We further include passive forensic references, including frame consistency, frequency analysis, and pixel statistics, as non-watermark baselines under the same evaluation protocol.

\textbf{Evaluation metrics.}
We evaluate LAVA in terms of detection accuracy, localisation quality, calibration reliability, and watermark fidelity.
Specifically, we use AP ($\uparrow$) and AUC ($\uparrow$) to measure tamper detection performance at the frame and video levels, respectively, and adopt temporal IoU ($\uparrow$) to evaluate localisation quality.
To assess the reliability of the predicted tamper probabilities, we further report ECE ($\downarrow$).
In addition, we use PSNR ($\uparrow$), SSIM ($\uparrow$), and FPR ($\downarrow$) to evaluate the perceptual quality and false-alarm behaviour of the embedded watermark.
The offset bank uses oracle selection (ground-truth labels) to establish the upper bound of alignment-aware fusion.
$T^*$ is tuned on a held-out validation split disjoint from the test set.

\begin{table*}[!t]
\centering
\caption{Ablation on LAV-DF ($n{=}500$). Each row adds one pipeline layer. Offset bank alone can \emph{degrade} AP under JPEG (row~2); the gate resolves this (row~3). Best in \textbf{bold}, second-best \underline{underlined}.}
\label{tab:ablation}
\small
\begin{tabular}{l ccc ccc ccc ccc}
\toprule
 & \multicolumn{3}{c}{Clean} & \multicolumn{3}{c}{JPEG $q{=}$23} & \multicolumn{3}{c}{H.264 CRF\,23} & \multicolumn{3}{c}{H.264 CRF\,28} \\
\cmidrule(lr){2-4} \cmidrule(lr){5-7} \cmidrule(lr){8-10} \cmidrule(lr){11-13}
 & AP & IoU & ECE & AP & IoU & ECE & AP & IoU & ECE & AP & IoU & ECE \\
\midrule
Na\"{\i}ve fusion & \textbf{1.000} & 0.950 & 0.006 & \underline{0.999} & \underline{0.924} & 0.441 & \textbf{1.000} & 0.873 & 0.419 & \textbf{1.000} & 0.800 & 0.420 \\
w/ Offset Bank    & \textbf{1.000} & 0.950 & 0.006 & 0.895 & 0.801 & 0.443 & -- & -- & -- & -- & -- & -- \\
w/ OB + Gate      & \textbf{1.000} & \underline{0.944} & \underline{0.003} & \underline{0.999} & \underline{0.924} & \underline{0.002} & \textbf{1.000} & \textbf{0.981} & \underline{0.004} & \textbf{1.000} & \textbf{0.981} & \underline{0.004} \\
LAVA (full)       & \underline{1.000} & \textbf{0.952} & \textbf{0.001} & \textbf{0.999} & \textbf{0.955} & \textbf{0.002} & \textbf{1.000} & \textbf{0.981} & \textbf{0.003} & \textbf{1.000} & \textbf{0.981} & \textbf{0.003} \\
\bottomrule
\end{tabular}
\end{table*}

\subsection{Overall Detection and Localisation Performance}
\label{sec:exp:main}

Table~\ref{tab:main} shows that LAVA delivers the most consistent overall performance across datasets and distortion conditions. Under clean settings, performance is already near saturation on LAV-DF, while LAVA still attains the best results on FakeAVCeleb and VoxCeleb2, reaching AUC\,=\,1.000 and AP\,=\,0.991, respectively. Under JPEG compression, its advantage becomes more pronounced: LAVA preserves AP\,=\,0.999 on LAV-DF and AUC\,=\,0.810 on FakeAVCeleb, and achieves the best localisation result on LAV-DF with IoU\,=\,0.955. Relative to na\"{\i}ve fusion, this corresponds to an improvement of 0.205 in JPEG IoU on LAV-DF and 0.018 in JPEG AUC on FakeAVCeleb.

The main difference from the comparison variants is that LAVA remains effective when the visual branch becomes unreliable. This is clearest under JPEG compression, where the visual-only baseline drops sharply on all three datasets, from 1.000 to 0.453 AP on LAV-DF, from 0.833 to 0.565 AUC on FakeAVCeleb, and from 0.690 to 0.214 AP on VoxCeleb2. By contrast, LAVA retains strong performance by suppressing corrupted visual evidence and falling back on the surviving audio cue when necessary. This effect is particularly evident on FakeAVCeleb, where LAVA reaches AUC\,=\,0.810, clearly outperforming visual-only (0.565) and na\"{\i}ve fusion (0.792). On VoxCeleb2 under JPEG compression, the audio-only variant is marginally stronger than LAVA in AP (0.780 vs.\ 0.773), suggesting that when discriminative evidence is already concentrated in the audio channel, the gain from cross-modal fusion becomes limited rather than uniformly positive.

\subsection{Component Analysis}
\label{sec:exp:ablation}

Table~\ref{tab:ablation} shows that the benefit of LAVA does not come from offset alignment alone. In fact, adding the offset bank without any reliability control already exposes a clear failure mode: although clean performance remains unchanged, JPEG performance drops noticeably, with AP decreasing from 0.999 to 0.895 and IoU from 0.924 to 0.801. This result suggests that, once the visual branch becomes unreliable, offset search can be misled by corrupted evidence rather than improving fusion.

The reliability gate is the component that resolves this failure. Relative to the offset-only variant, adding the gate restores JPEG AP from 0.895 to 0.999 and IoU from 0.801 to 0.924. The same effect becomes even clearer under stronger compression: at H.264 CRF\,23 and CRF\,28, IoU increases from 0.873 and 0.800 to 0.981 and 0.981, while ECE drops from 0.419 and 0.420 to 0.004 and 0.004, respectively. This confirms that the main role of the gate is to prevent corrupted visual evidence from dominating the fused prediction.

The remaining gain comes from completing the full pipeline. After confidence-weighted fusion and calibration are added, the model reaches the best clean IoU/ECE of 0.952/0.001 and the best JPEG IoU of 0.955, while keeping ECE at or below 0.003 across all four compression settings. The ablation therefore points to a clear division of labor: offset alignment restores temporal consistency, the gate handles modality failure, and the full pipeline is needed to obtain the final gains in localisation quality and calibration.

\begin{table}[h]
\centering
\small
\caption{Temporal robustness on LAV-DF ($n{=}500$). (a)~LAVA recovers AP\,$=$\,1.000 for all in-bank offsets. (b)~Linear resampling recovers AP\,$>$\,0.999.}
\label{tab:offset_stretch}
\setlength{\tabcolsep}{4pt}
\begin{tabular}{r c rr r}
\toprule
\multicolumn{5}{c}{\textit{(a) A/V Offset Stress}} \\
Offset & In bank & Na\"{\i}ve & LAVA & $\Delta$ \\
\midrule
+0.0\,s & \cmark & 1.000 & \textbf{1.000} & +0.000 \\
+0.5\,s & \cmark & 0.818 & \textbf{1.000} & +0.182 \\
+1.0\,s & \cmark & 0.640 & \textbf{1.000} & +0.360 \\
+2.0\,s & \xmark & 0.779 & \textbf{0.912} & +0.133 \\
+3.0\,s & \xmark & 0.903 & \textbf{0.961} & +0.058 \\
\midrule
\multicolumn{5}{c}{\textit{(b) Audio Stretch Correction}} \\
Factor & & Raw & Corrected & Recovery \\
\midrule
0.90$\times$ & & 0.132 & \textbf{0.998} & 99.9\% \\
0.95$\times$ & & 0.131 & \textbf{0.999} & 100.0\% \\
1.05$\times$ & & 0.148 & \textbf{0.999} & 100.0\% \\
1.10$\times$ & & 0.903 & \textbf{0.999} & 99.4\% \\
\bottomrule
\end{tabular}
\end{table}

\subsection{Robustness Analysis}
\label{sec:exp:robustness}

To verify the robustness of LAVA under temporal misalignment, tempo perturbation, and joint cross-modal degradation, we report results on LAV-DF and VoxCeleb2 in Tables~\ref{tab:offset_stretch} and~\ref{tab:joint_vox}. The results show that LAVA remains highly robust to moderate temporal distortions and consistently outperforms na\"{\i}ve fusion under realistic multi-factor degradations.

A closer examination reveals three main findings. First, LAVA effectively handles temporal offsets within the predefined offset bank. Under offsets of $+0.5$\,s and $+1.0$\,s, na\"{\i}ve fusion drops to AP\,=\,0.818 and 0.640, respectively, whereas LAVA fully recovers AP\,=\,1.000 in both cases. Even for out-of-bank offsets, LAVA still improves AP from 0.779 to 0.912 at $+2.0$\,s and from 0.903 to 0.961 at $+3.0$\,s, indicating that offset-aware fusion remains beneficial beyond the nominal search range.

Second, audio stretch severely disrupts the raw audio watermark signal, but the proposed correction largely restores performance. Specifically, the raw AP drops to 0.132, 0.131, and 0.148 under stretch factors of 0.90$\times$, 0.95$\times$, and 1.05$\times$, respectively. After linear-resampling correction, the AP is restored to 0.998--0.999, corresponding to recovery rates of 99.9\%--100.0\%. Even at 1.10$\times$, where the raw AP is already 0.903, correction further improves it to 0.999.

Third, the advantage of LAVA becomes more pronounced under joint degradations. On LAV-DF with an injected $+0.5$\,s offset, LAVA consistently outperforms na\"{\i}ve fusion across all visual/audio combinations. The largest gain appears under JPEG + MP3 32\,k, where AP increases from 0.398 to 0.982. Strong improvements are also observed under JPEG + Clean (0.456$\to$0.989) and Clean + MP3 32\,k (0.842$\to$1.000), showing that the layered design remains effective when one modality is substantially degraded. The most challenging case is JPEG + Stretch 0.95$\times$, where LAVA still improves AP from 0.284 to 0.504, but simultaneous degradation of both channels leads to a clear performance drop. On VoxCeleb2, LAVA remains robust across diverse visual attack types, achieving fused APs of 0.962--1.000 under clean conditions and 0.760--0.788 under JPEG compression. Taken together, these results show that LAVA is robust not only to isolated temporal distortions, but also to realistic joint cross-modal degradations.

\begin{table}[t]
\centering
\small
\caption{Joint A/V attacks. (a)~LAV-DF with $+$0.5\,s offset. (b)~VoxCeleb2 async 0.65\,s tamper. LAVA gains up to $+$0.584. AP values shown.}
\label{tab:joint_vox}
\setlength{\tabcolsep}{4pt}
\begin{tabular}{ll rr r}
\toprule
\multicolumn{5}{c}{\textit{(a) Joint Attacks (LAV-DF)}} \\
Visual & Audio & Na\"{\i}ve & LAVA & $\Delta$ \\
\midrule
Clean & Clean & 0.818 & \textbf{1.000} & +0.182 \\
Clean & MP3\,32\,k & 0.842 & \textbf{1.000} & +0.158 \\
Clean & Stretch\,0.95$\times$ & 0.968 & \textbf{1.000} & +0.032 \\
JPEG & Clean & 0.456 & \textbf{0.989} & +0.533 \\
JPEG & MP3\,32\,k & 0.398 & \textbf{0.982} & +0.584 \\
JPEG & Stretch\,0.95$\times$ & 0.284 & \textbf{0.504} & +0.220 \\
\midrule
\multicolumn{5}{c}{\textit{(b) Visual Attacks (VoxCeleb2)}} \\
\multicolumn{2}{l}{Attack} & \multicolumn{2}{c}{Fused AP} & \\
\cmidrule(lr){3-4}
\multicolumn{2}{l}{} & Clean & JPEG & \\
\midrule
\multicolumn{2}{l}{Blur} & \textbf{0.962} & \textbf{0.777} & \\
\multicolumn{2}{l}{Black-box} & \textbf{1.000} & \textbf{0.768} & \\
\multicolumn{2}{l}{Mosaic} & \textbf{1.000} & \textbf{0.788} & \\
\multicolumn{2}{l}{Noise} & \textbf{1.000} & \textbf{0.774} & \\
\multicolumn{2}{l}{White-out} & \textbf{1.000} & \textbf{0.760} & \\
\bottomrule
\end{tabular}
\end{table}

\subsection{Spatiotemporal Detection}
\label{sec:exp:spatial}

\begin{figure}[!t]
\centering
\includegraphics[width=0.49\textwidth]{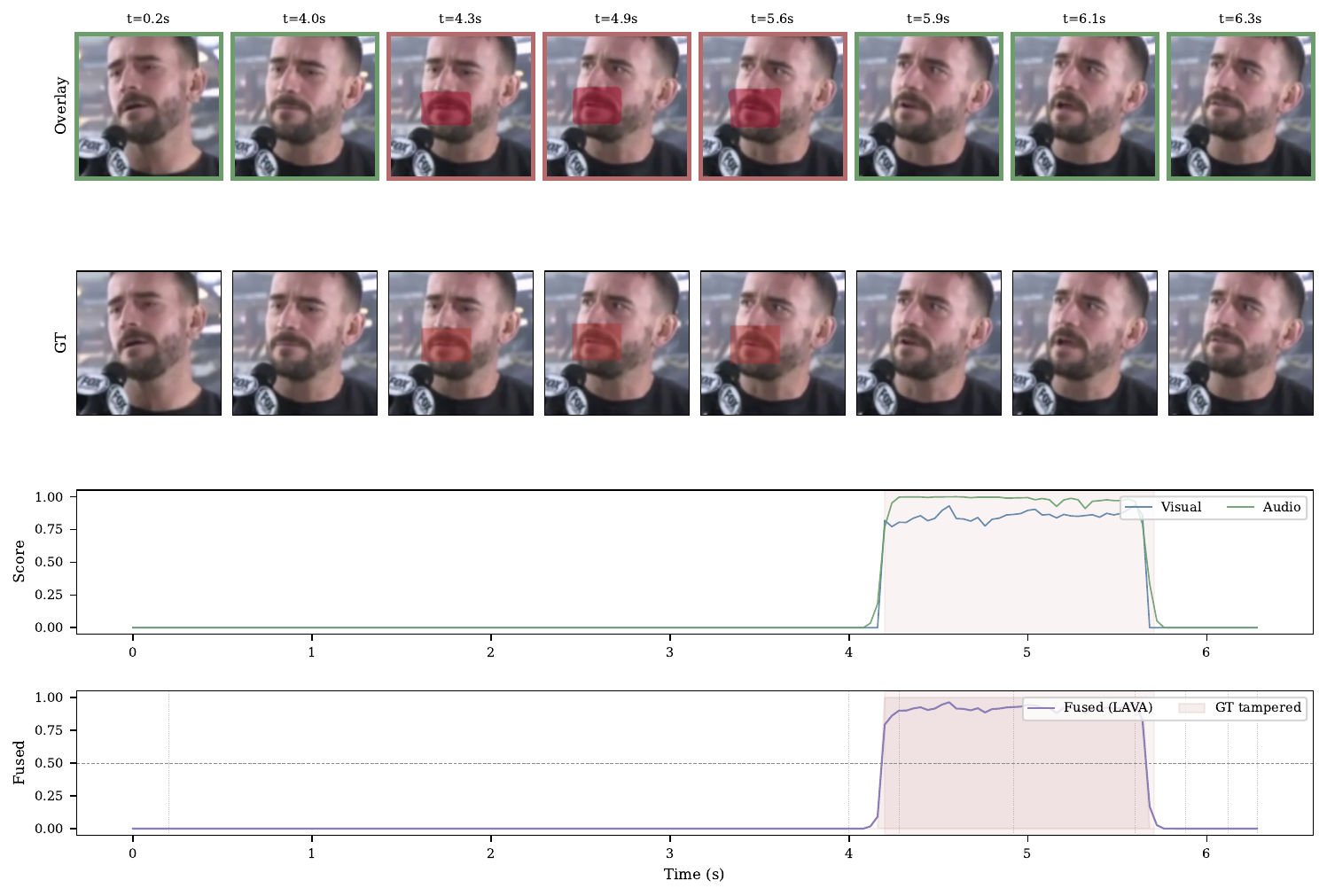}
\Description{Top rows show per-frame tamper heatmap overlays on video frames sampled across time, with green borders for authentic frames and red borders for tampered frames. Bottom rows show temporal score traces for visual, audio, and LAVA fused signals with ground-truth tampered intervals shaded in red.}
\caption{Spatiotemporal detection on LAV-DF.
\textbf{Top}: per-frame heatmap overlays across time
(green\,$=$\,authentic, red\,$=$\,tampered).
\textbf{Bottom}: visual, audio, and LAVA fused score traces with GT tampered intervals shaded.}
\label{fig:spatiotemporal}
\end{figure}

Figure~\ref{fig:spatiotemporal} makes the behavior of LAVA concrete on a representative LAV-DF example. The temporal trace remains tightly aligned with the ground-truth tampered interval, with clear transitions at both boundaries, while the spatial response is concentrated around the manipulated mouth region rather than spreading over the full face. Authentic frames stay near the background level throughout, whereas tampered frames trigger localized and temporally coherent activations.

Table~\ref{tab:loc_attr} quantifies this accuracy on raw visual maps \emph{before} gate-based abstention. Under clean conditions, the visual detector reaches IoU\,=\,0.963 and F1\,=\,0.981, indicating close agreement between the predicted tamper masks and the ground truth. JPEG compression makes this task markedly harder by enlarging the activated regions and reducing spatial precision. Even in this regime, a simple morphological closing step improves IoU from 0.132 to 0.187 and F1 from 0.233 to 0.315, suggesting that the degradation is partly geometric, i.e., fragmented or distorted masks, rather than a complete disappearance of spatial evidence. Spatial localization therefore remains meaningful when the visual watermark is preserved, and deteriorates mainly in the heavily compressed setting.

\subsection{Attribution, Passive Comparison, and Watermark Quality}
\label{sec:exp:attr}

Table~\ref{tab:loc_attr} extends the evaluation beyond binary tamper detection to three complementary aspects: manipulation-type attribution, comparison with passive forensic cues, and watermark fidelity. Two patterns are immediately clear. The first is that the watermark responses remain informative enough to support attribution without any additional classifier. The second is that this additional functionality is obtained without sacrificing imperceptibility or increasing false alarms.

For manipulation-type attribution, the rule-based scheme is highly accurate under clean conditions. It reaches 0.995 accuracy on LAV-DF and 1.000 on FakeAVCeleb, indicating that the joint audio-visual watermark responses are sufficiently structured to separate authentic content, face swap, voice cloning, and joint deepfake cases. This capability weakens under JPEG compression, where FakeAVCeleb accuracy drops to 0.500, consistent with the loss of discriminative visual evidence needed for fine-grained type inference.

The passive baselines remain far behind in the same setting. On LAV-DF, frame consistency, frequency analysis, and pixel statistics achieve APs of only 0.042, 0.090, and 0.152, respectively, whereas LAVA reaches AP\,=\,0.999. This gap suggests that the manipulated videos are realistic enough to suppress conventional low-level forensic artifacts, while watermark absence remains a much more direct and reliable cue for tamper detection.

The watermark itself also remains highly unobtrusive. On VoxCeleb2, the visual branch achieves PSNR\,=\,39.2\,dB and SSIM\,=\,0.999, while the audio branch reaches SNR\,=\,25.3\,dB. The false positive rate is 0.000 for both modalities, so authentic watermarked content is never mistakenly flagged as tampered. In other words, the same embedded signal that supports attribution and detection introduces almost no perceptual burden in practice.

\begin{table}[t]
\centering
\small
\caption{Spatial localisation (LAV-DF), cross-modal attribution (LAV-DF + FakeAVCeleb), passive baselines (LAV-DF), and watermark quality (VoxCeleb2). LAVA achieves 0.995--1.000 attribution accuracy and zero false positives.}
\label{tab:loc_attr}
\setlength{\tabcolsep}{4pt}
\begin{tabular}{l l ccc}
\toprule
\multicolumn{5}{c}{\textit{(a) Spatial Localisation (LAV-DF, 5 groups)}} \\
Condition & Variant & IoU & Recall & F1 \\
\midrule
Clean & Baseline & \textbf{0.963} & 0.978 & \textbf{0.981} \\
Clean & Refined  & \textbf{0.963} & 0.978 & \textbf{0.981} \\
JPEG  & Baseline & 0.132 & \textbf{0.995} & 0.233 \\
JPEG  & Refined  & \textbf{0.187} & 0.972 & \textbf{0.315} \\
\midrule
\multicolumn{5}{c}{\textit{(b) Cross-Modal Attribution}} \\
Dataset & Condition & \multicolumn{3}{c}{Accuracy} \\
\midrule
LAV-DF & Clean & \multicolumn{3}{c}{\textbf{0.995}} \\
FakeAVCeleb & Clean & \multicolumn{3}{c}{\textbf{1.000}} \\
FakeAVCeleb & JPEG & \multicolumn{3}{c}{0.500} \\
\midrule
\multicolumn{5}{c}{\textit{(c) Passive Baselines (LAV-DF, 50 groups)}} \\
Method & & \multicolumn{3}{c}{AP} \\
\midrule
Frame consistency & & \multicolumn{3}{c}{0.042} \\
Frequency analysis & & \multicolumn{3}{c}{0.090} \\
Pixel statistics & & \multicolumn{3}{c}{0.152} \\
\textbf{LAVA (ours)} & & \multicolumn{3}{c}{\textbf{0.999}} \\
\midrule
\multicolumn{5}{c}{\textit{(d) Watermark Quality (VoxCeleb2, 200 videos)}} \\
Metric & & Visual & \multicolumn{2}{c}{Audio} \\
\midrule
PSNR / SNR & & 39.2\,dB & \multicolumn{2}{c}{25.3\,dB} \\
SSIM & & 0.999 & \multicolumn{2}{c}{---} \\
FPR & & 0.000 & \multicolumn{2}{c}{0.000} \\
\bottomrule
\end{tabular}
\end{table}

\subsection{Comparison with External Watermark Methods}
\label{sec:exp:external}

To further contextualise LAVA's performance, we compare it against two external active watermark methods evaluated under the same protocol on LAV-DF: \textbf{EditGuard}~\cite{Zhang_2024_CVPR}, a semi-fragile visual watermark designed for image tamper localisation, and \textbf{WavMark}~\cite{chen2024wavmarkwatermarkingaudiogeneration}, a neural audio watermark that embeds a 32-bit payload into 16\,kHz speech.
Both methods follow LAVA's evaluation pipeline: watermarks are first embedded into the original media, the composite is then constructed by replacing tampered segments with unwatermarked deepfake content according to LAV-DF's ground-truth intervals, and detection is performed on the resulting composite.
We additionally include two learned passive detectors, XceptionNet~\cite{Rossler_2019_ICCV++} and EfficientNet-B4~\cite{tan2020efficientnetrethinkingmodelscaling}, both pretrained on FaceForensics++~\cite{Rossler_2019_ICCV++} (c23) and evaluated in a zero-shot cross-dataset setting.

Table~\ref{tab:external} reports frame-level AP under clean conditions.
Among the external methods, EditGuard performs best with AP\,=\,0.549, confirming that proactive watermarking substantially outperforms passive detection (XceptionNet AP\,=\,0.039, EfficientNet-B4 AP\,=\,0.050).
However, EditGuard's performance is highly variable across videos (per-group AP\,=\,0.602\,$\pm$\,0.356), with some groups achieving perfect detection while others fall below 10\%, suggesting that its semi-fragile design does not transfer reliably to mouth-region deepfakes.
WavMark achieves AP\,=\,0.354 on audio tampering, indicating that voice cloning largely destroys the embedded payload, limiting its utility as a standalone tamper detector.

LAVA outperforms all external methods by a wide margin.
The key advantage is \emph{cross-modal complementarity}: EditGuard and WavMark each operate on a single modality and fail when that modality's watermark is disrupted by the forgery process.
By contrast, LAVA fuses visual and audio watermark evidence through reliability-aware gating and temporal alignment, achieving AP\,$\geq$\,0.999 under both clean and compressed conditions.
This comparison highlights that the performance gap is not merely due to a stronger base watermark, but arises from LAVA's layered multi-modal fusion design.

\begin{table}[t]
\centering
\small
\caption{Comparison with external methods on LAV-DF (clean).
Active watermark methods embed their respective watermark into original media before constructing the tampered composite.
Passive detectors are applied zero-shot (trained on FF++ c23).
AP\,=\,frame-level average precision.}
\label{tab:external}
\setlength{\tabcolsep}{4pt}
\begin{tabular}{l l c c}
\toprule
Method & Type & Modality & AP \\
\midrule
\multicolumn{4}{l}{\emph{Active watermark}} \\
EditGuard~\cite{Zhang_2024_CVPR}    & Semi-fragile WM     & Visual & 0.549 \\
WavMark~\cite{chen2024wavmarkwatermarkingaudiogeneration}        & Neural audio WM     & Audio  & 0.354 \\
Visual-only (WAM)            & Semi-fragile WM     & Visual & \underline{1.000} \\
Audio-only (AudioSeal)       & Semi-fragile WM     & Audio  & \underline{0.999} \\
\midrule
\multicolumn{4}{l}{\emph{Passive detector (zero-shot, FF++\,$\to$\,LAV-DF)}} \\
XceptionNet~\cite{Rossler_2019_ICCV++}     & Learned & Visual & 0.039 \\
EfficientNet-B4~\cite{tan2020efficientnetrethinkingmodelscaling}      & Learned & Visual & 0.050 \\
\midrule
\textbf{LAVA (ours)} & Multi-modal WM & A\,+\,V & \textbf{1.000} \\
\bottomrule
\end{tabular}
\end{table}

\section{Conclusion}
\label{sec:conclusion}

We presented LAVA, a layered audio-visual anti-tampering framework
that fuses independent visual and audio semi-fragile watermarks
through temporal stretch correction, reliability gating,
confidence-weighted fusion, and calibration.
Experiments on LAV-DF, FakeAVCeleb, and VoxCeleb2 demonstrate
near-perfect detection (AP\,$\geq$\,0.999), robust localisation
(IoU\,$=$\,0.955), well-calibrated outputs (ECE\,$\leq$\,0.002),
and 99.5--100\% attribution accuracy---all with imperceptible
watermarks (PSNR\,$=$\,39.2\,dB, SSIM\,$=$\,0.999) and zero false
positives. Under H.264 CRF\,28 compression, the reliability gate
preserves IoU\,$=$\,0.981 ($+$18\,pp over na\"{\i}ve fusion).
The current design assumes temporally sparse tampering and relies on
oracle offset selection; future work includes learned alignment,
adversarial robustness, and C2PA integration.

\bibliographystyle{ACM-Reference-Format}
\bibliography{ref}

\end{document}